\documentclass[runningheads]{llncs}
\usepackage[T1]{fontenc}
\usepackage{graphicx}
\usepackage{amsmath}
\usepackage{booktabs}
\usepackage{amsfonts}
\usepackage{amsmath,graphicx,hyperref}
\usepackage{caption}
\usepackage{booktabs}
\usepackage{multirow}
\usepackage{xcolor}
\usepackage{tabularx}
\usepackage{placeins}
\begin{document}
\title{PASTA: Prompt-Aware Shared-Latent Training-Free Annealed Diffusion Sampling}

\titlerunning{PASTA: Prompt-Aware Annealed Diffusion Sampling}
\author{
Ren Ajiki\inst{1}\textsuperscript{*} \and
Ryugo Morita\inst{2}\textsuperscript{*} \and
Stanislav Frolov\inst{2} \and
Brian Bernhard Moser\inst{2}  \\
Ko Watanabe\inst{2} \and
Riku Takahashi\inst{1} \and
Andreas Dengel\inst{2} \and
Jinjia Zhou\inst{1}
}

\authorrunning{R. Ajiki et al.}

\institute{
Faculty of Science and Engineering, Hosei University, Tokyo, Japan
\and
RPTU Kaiserslautern-Landau \& DFKI GmbH, Kaiserslautern, Germany
}

\maketitle              %
\begingroup
\renewcommand{\thefootnote}{*}
\footnotetext{Equal contribution.}
\endgroup
\begin{abstract}
While Diffusion models produce high-quality images, real-world visual design requires intentionally preserving clean regions for headlines, logos, or UI elements.
Existing layout-aware text-to-image methods struggle to strictly preserve reserved regions, often introducing unwanted content that complicates downstream editing.
Existing approaches either manipulate attention to displace content or suppress conditioning via noise-level interventions.
However, these methods sacrifice visual coherence, producing flat regions or noticeable style inconsistencies between reserved and non-reserved areas.
To address this limitation, we propose a Prompt-Aware Shared-Latent Training-Free Annealed Diffusion Sampling (PASTA), a training-free framework that enforces reserved-region constraints by selectively applying semantic guidance to object-allowed regions during sampling.
PASTA operates on a shared latent trajectory, where full-prompt and context-prompt guidance are fused through mask-guided sampling. An annealed schedule suppresses semantic leakage during early denoising stages while gradually restoring visual consistency in later stages.
Extensive experiments demonstrate that PASTA improves reserved-region fidelity while preserving overall image realism and perceptual consistency.

\keywords{Space-Aware Text-to-Image, Image Generation, Diffusion Models}
\end{abstract}
\begin{figure}[t]
    \centering
    \includegraphics[width=\linewidth]{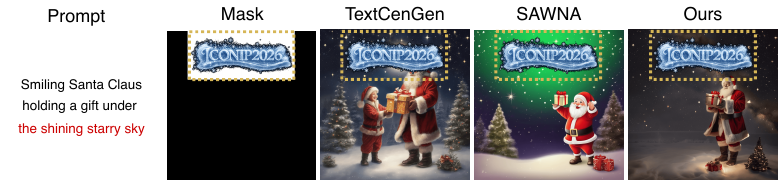}
    \caption{
Red words in the prompt column indicate the context prompt \(P_{\mathrm{ctx}}\), and yellow dashed boxes denote the reserved regions.
TextCenGen~\cite{liang2024textcengen}  causes foreground content to overlap with the placed logo-style text element, while SAWNA~\cite{morita2025sawna}  produces a flat and visually inconsistent reserved region.
PASTA keeps the reserved region clean while maintaining a coherent background.
}
    \label{fig:teaser}
\end{figure}

\section{Introduction}
\label{sec:introduction}
Diffusion models~\cite{rombach2022high,peebles2023scalable} have enabled high-fidelity text-to-image generation from text prompts, reshaping visual content creation in art~\cite{li2024adversarial,koley2024s}, design~\cite{wang2025designdiffusion,gao2025postermaker}, and media~\cite{meng2025anidoc,liu2025manganinja}.
In real-world design, however, creators often need not only to place objects but also to reserve clean regions for text, logos, or interface elements, such as headlines, UI components, or subtitles.
These reserved regions should remain free of salient content while visually matching the surrounding background.
Existing layout-aware text-to-image methods do not treat reserved regions as first-class constraints, leading to unwanted objects or distracting textures that complicate downstream editing.

Several recent works have explored space-aware text-to-image generation to address this need.
TextCenGen~\cite{liang2024textcengen} manipulates cross-attention to push object evidence away from a target region and free space for text placement.
While effective when only mild repositioning is needed, attention relocation may have difficulty satisfying large, strictly reserved regions.
SAWNA~\cite{morita2025sawna} injects non-reactive noise into reserved regions, ensuring that no objects are generated from the start.
Although this strategy reliably enforces empty space, suppression produces monotone regions, leading to noticeable visual inconsistencies between reserved and non-reserved areas. As illustrated in Fig.~\ref{fig:teaser}, existing methods either allow foreground content to intrude into the reserved region or produce visually inconsistent backgrounds.

To address this gap, we propose a novel Prompt-Aware Shared-Latent Training-Free Annealed Diffusion Sampling (PASTA) that explicitly enforces reserved-region constraints while preserving background naturalness via annealed mask-guided fusion of full-prompt and context-prompt guidance.
At each diffusion step, PASTA evaluates a shared latent under negative-prompt, full-prompt, and context-prompt conditions, derives guidance directions relative to the negative-prompt prediction, and spatially fuses them using a user-defined reserved-region mask.
An annealed blending schedule further prevents early semantic leakage while enabling natural refinement in later steps.

Extensive experiments demonstrate that PASTA improves fidelity in reserved regions while maintaining competitive performance in both quantitative and perceptual evaluations.
Our contributions are summarized as follows:
\begin{itemize}
\item We propose PASTA, a prompt-aware shared-latent training-free diffusion sampling framework that spatially fuses guidance residuals to suppress foreground semantic leakage in reserved regions while preserving generation consistency in editable regions.

\item We introduce an annealed guidance blending schedule that applies stronger context-prompt guidance during early denoising stages and progressively restores full-prompt guidance in later stages, improving reserved-region fidelity and overall image coherence.

\item We conduct extensive quantitative and qualitative evaluations across diverse generation settings, demonstrating that PASTA achieves improved reserved-region fidelity while maintaining competitive overall visual quality.
\end{itemize}

\section{Related Work}
\label{sec:related_work}

\subsection{Controllable Diffusion-based Image Generation}
Diffusion models have become a dominant framework for high-fidelity image synthesis by iteratively denoising Gaussian noise, and large-scale text-conditioned variants enable photorealistic and diverse generation from prompts~\cite{rombach2022high}.
To improve controllability, 
prior work has explored various conditioning strategies, including structural cues (edges, sketches)~\cite{liu2025sketchvideo,ding2024training}, geometric cues (depth, pose)~\cite{xiao2025omnigen,zhang2023adding}, and low-level appearance constraints (color)~\cite{cohen2026colorful,pach2026latent}.
Moreover, training-free inference-time control methods manipulate guidance~\cite{ho2022classifier,chung2025cfg++}, attention~\cite{epstein2023diffusion,nagai2025taue}, or latent representations~\cite{morita2025tkg,morita2026lgtm} to achieve flexible generation behavior without additional training.
Recent studies have further explored layered representations that decompose images into editable foreground and background components, aligning generation with designer-centric pipelines~\cite{yin2025qwen,huang2025dreamlayer}.
As layered generation matures, designers increasingly compose visuals by editing separate layers for text and foreground objects.
This workflow introduces a new requirement for text-to-image generation: preserving clean reserved regions that remain free from unintended foreground semantics during generation.

\subsection{Spatially-aware Text-to-Image Generation}
Layout-aware text-to-image generation improves compositional controllability by conditioning generation on spatial constraints such as bounding boxes or region masks~\cite{li2023gligen,yang2023reco}.
Existing methods achieve spatial control either through explicit conditioning mechanisms or by manipulating attention and guidance during sampling in a training-free manner~\cite{bar2023multidiffusion,xie2023boxdiff}. These methods mainly use bounding boxes or region masks to specify where target objects should appear in the generated image.

Our setting is fundamentally different.  
Instead of attracting foreground content into designated regions, the objective of reserved-region generation is to prevent salient foreground semantics from occupying reserved regions.
Although layout-guided placement methods are related to spatial controllability, these methods are not direct baselines for the reserved-region objective considered in this work.
In real design workflows, however, specifying object locations is often insufficient.
Designers frequently need to reserve clean regions for text, logos, or interface elements, which are typically added or edited outside the generative model.

To address this problem, space-aware text-to-image generation has been explored to explicitly enforce empty regions during sampling.
TextCenGen~\cite{liang2024textcengen} relocates object-related attention away from regions reserved for text placement, but may have difficulty satisfying large and strictly reserved regions.
SAWNA~\cite{morita2025sawna} instead suppresses generation from the start via non-reactive noise, preserving space at the cost of producing flat or visually inconsistent reserved regions.

Motivated by these limitations, we propose a training-free sampling strategy that enforces clean reserved regions while preserving background naturalness via mask-guided fusion of full-prompt/context-prompt guidance with an annealed schedule.

\section{PASTA Framework}
\label{sec:method}

\begin{figure*}[t]
    \centering
    \includegraphics[width=\linewidth]{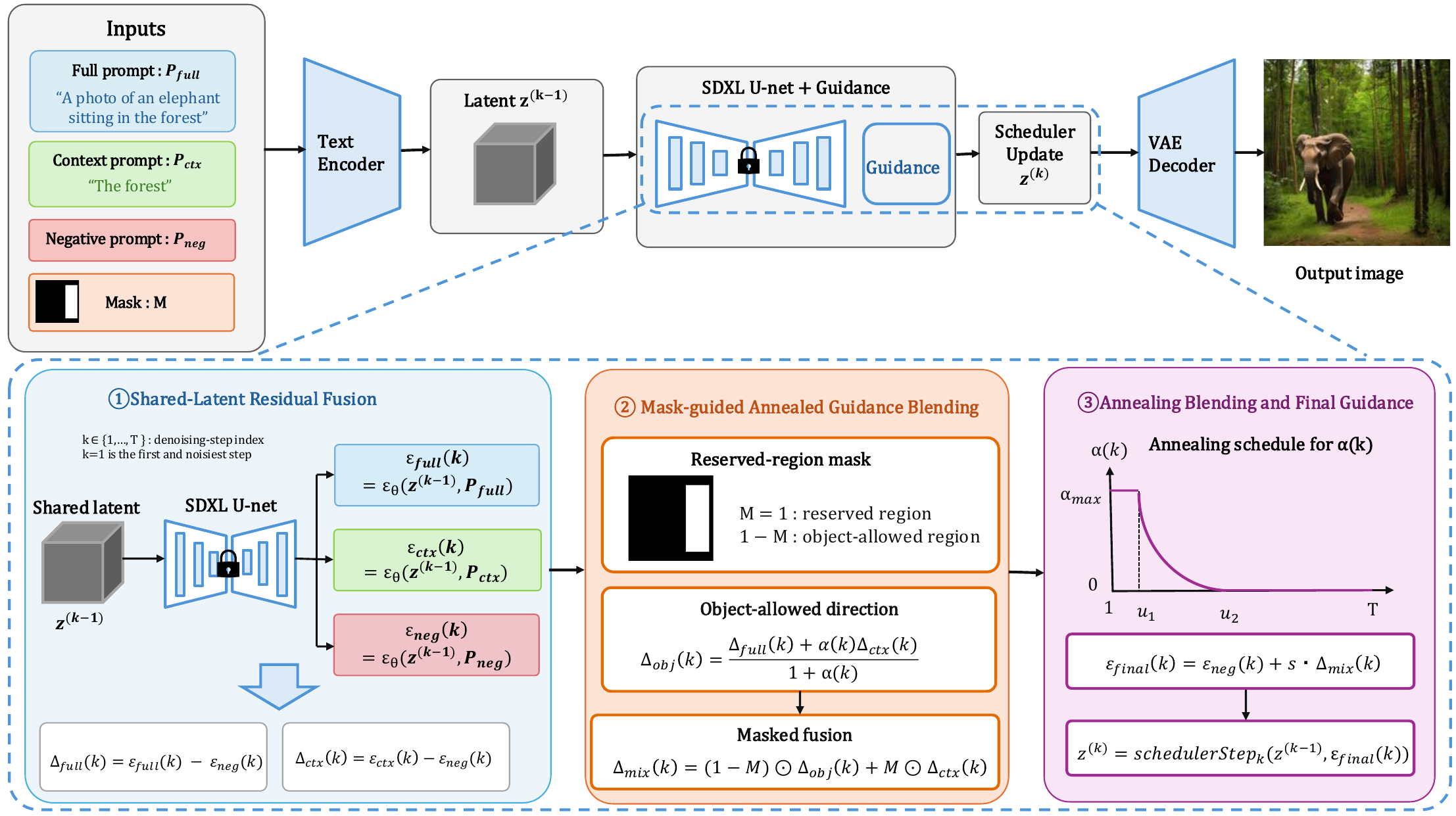}
\caption{Overview of PASTA. Given a full prompt $P_{\mathrm{full}}$ and a
reserved-region mask $M$, PASTA computes guidance residuals from shared-latent
noise predictions under full-prompt, context-prompt, and negative-prompt
conditions. The residuals are spatially fused to suppress foreground-object
guidance inside the reserved region, while an annealed schedule controls the
context-prompt contribution in the object-allowed region.}
    \label{fig:model}
\end{figure*}

As shown in Fig.~\ref{fig:model}, PASTA is a training-free diffusion sampling framework
that spatially fuses shared-latent guidance residuals
to suppress foreground semantic leakage inside reserved regions. Given a full prompt \(P_{\mathrm{full}}\) and a reserved-region mask \(M \in \{0,1\}^{h \times w}\) at the latent resolution, where \(M=1\) denotes the region to keep free of salient foreground content, PASTA modifies the classifier-free guidance direction during sampling.

We construct a context-prompt guidance \(P_{\mathrm{ctx}}\) by removing the main-object descriptors from \(P_{\mathrm{full}}\) while preserving contextual descriptors such as background, scene, and style.
Specifically, we use ChatGPT~5.2 with temperature \(0\) and a fixed decomposition template to split \(P_{\mathrm{full}}\) into a \texttt{foreground} phrase and a \texttt{context} phrase, and set \(P_{\mathrm{ctx}}\) to the \texttt{context} phrase.
Thus, \(P_{\mathrm{ctx}}\) provides context guidance without explicitly encouraging the foreground object that should be excluded from the reserved region.

\subsection{Shared-Latent Residual Fusion}
\label{sec:tripass_fusion}

Let \(T\) denote the total number of inference steps, and let \(k \in \{1,\ldots,T\}\) denote the index of the denoising step along the reverse sampling trajectory.
Here, \(k=1\) corresponds to the first and noisiest denoising step, and \(k=T\) corresponds to the final denoising step.
We denote by \(z^{(k-1)}\) the latent input to the \(k\)-th denoising step.

At each denoising step \(k\), PASTA evaluates the same UNet on the same latent \(z^{(k-1)}\) under three conditions: negative-prompt, full-prompt-conditioned, and context-prompt-conditioned.
Formally, we compute
\begin{equation}
\epsilon_{\mathrm{neg}}(k) = \epsilon_\theta(z^{(k-1)}, P_{\mathrm{neg}}), \quad
\epsilon_{\mathrm{full}}(k) = \epsilon_\theta(z^{(k-1)}, P_{\mathrm{full}}), \quad
\epsilon_{\mathrm{ctx}}(k) = \epsilon_\theta(z^{(k-1)}, P_{\mathrm{ctx}}),
\label{eq:tripass}
\end{equation}
where \(P_{\mathrm{neg}}\) denotes the negative prompt used in classifier-free guidance.

Using the negative-prompt prediction as the baseline, we define the guidance residuals:
\begin{equation}
\Delta_{\mathrm{full}}(k) = \epsilon_{\mathrm{full}}(k) - \epsilon_{\mathrm{neg}}(k), \qquad
\Delta_{\mathrm{ctx}}(k) = \epsilon_{\mathrm{ctx}}(k) - \epsilon_{\mathrm{neg}}(k).
\label{eq:delta}
\end{equation}

The residual ($\Delta_{\mathrm{full}}(k)$) captures foreground-aware guidance induced by the full prompt, while ($\Delta_{\mathrm{ctx}}(k)$) represents context-prompt guidance obtained after removing foreground-related semantics.
Because both residuals are computed from the same shared latent state ($z^{(k-1)}$), they remain spatially aligned and can therefore be fused within a single denoising step to selectively suppress foreground semantic leakage inside reserved regions.
\subsection{Mask-guided Annealed Guidance Blending}
\label{sec:anneal_update}

PASTA applies different guidance directions inside and outside the reserved region to selectively suppress foreground semantic leakage while preserving object generation quality in editable regions.
Inside the reserved region, we use \(\Delta_{\mathrm{ctx}}(k)\) to suppress foreground-object semantics while retaining contextual guidance.
Outside the reserved region, the full-prompt guidance should be preserved to generate the foreground object.
However, strong full-prompt guidance in early denoising steps may cause object semantics to leak into the reserved region, especially under large reserved masks.

To control this trade-off, we introduce an annealed mixing weight \(\alpha(k)\).
We set the schedule boundaries as fixed fractions of the total number of inference steps:
\(u_1 = 0.1T\) and \(u_2 = 0.3T\).
The weight is kept at its maximum value during the initial denoising stage, decayed in the middle stage, and set to zero afterwards:
\begin{equation}
\alpha(k)=
\begin{cases}
\alpha_{\max}, & 1 \le k \le u_1,\\
\alpha_{\max}\left(1-\dfrac{k-u_1}{u_2-u_1}\right)^p, & u_1 < k \le u_2,\\
0, & u_2 < k \le T,
\end{cases}
\label{eq:alpha_piecewise}
\end{equation}
where \(\alpha_{\max}\) is the maximum mixing weight and \(p\) controls how rapidly the weight decays to zero.
This schedule applies stronger context-prompt guidance in the early denoising steps to suppress semantic leakage into the reserved region, and gradually restores full-prompt guidance in later steps to refine the foreground object.

In the object-allowed region, we combine the full-prompt residual and the context-prompt residual using a normalized weighted average:
\begin{equation}
\Delta_{\mathrm{obj}}(k)
=
\frac{\Delta_{\mathrm{full}}(k) + \alpha(k)\,\Delta_{\mathrm{ctx}}(k)}{1+\alpha(k)}.
\label{eq:weight_norm}
\end{equation}
This normalized form adjusts the relative contribution of \(\Delta_{\mathrm{full}}(k)\) and \(\Delta_{\mathrm{ctx}}(k)\) while avoiding the trivial increase in guidance magnitude that would occur when simply adding \(\alpha(k)\Delta_{\mathrm{ctx}}(k)\) to \(\Delta_{\mathrm{full}}(k)\).
Thus, \(\alpha(k)\) mainly controls the mixing ratio between full-prompt guidance and context-prompt guidance, rather than acting as an additional guidance-scale parameter.

The final fused guidance residual is computed as
\begin{equation}
\Delta_{\mathrm{mix}}(k)
=
(1-M)\odot \Delta_{\mathrm{obj}}(k)
+
M \odot \Delta_{\mathrm{ctx}}(k),
\label{eq:masked_fusion}
\end{equation}
where \(\odot\) denotes element-wise multiplication.
This fusion enables spatially selective guidance control by suppressing foreground-aware residuals inside reserved regions while preserving full-prompt generation behavior in editable regions.

Finally, we form the noise prediction with a single guidance scale \(s\):
\begin{equation}
\epsilon_{\mathrm{final}}(k) = \epsilon_{\mathrm{neg}}(k) + s\cdot \Delta_{\mathrm{mix}}(k),
\label{eq:eps_final}
\end{equation}
and update the latent using the scheduler step:
\begin{equation}
z^{(k)} = \mathrm{SchedulerStep}_{k}\!\left(z^{(k-1)}, \epsilon_{\mathrm{final}}(k)\right).
\label{eq:scheduler}
\end{equation}
The updated latent \(z^{(k)}\) is shared by all three prompt conditions at the next denoising step.

By maintaining a shared latent trajectory across all prompt conditions, PASTA avoids multi-trajectory inconsistencies and enables stable residual blending throughout the denoising process.

\section{Experiments}
\subsection{Experimental Setup}
\label{sec:setup}
We adopt Stable Diffusion XL (SDXL) in PASTA to generate images at $1024\times1024$ resolution.
We use UniPCMultistepScheduler with CFG scale \(s=6.5\) and \(T=50\) inference steps.
For annealing, we use Eq.~\eqref{eq:alpha_piecewise} with \(\alpha_{\max}=3\), \(p=2\), \(u_1=0.1T\), and \(u_2=0.3T\).
With \(T=50\), this corresponds to \(u_1=5\) and \(u_2=15\).
These hyperparameters were selected empirically in preliminary experiments and fixed for all reported results.
We compare our method with space-aware text-to-image methods, TextCenGen~\cite{liang2024textcengen} and SAWNA~\cite{morita2025sawna}. All baselines are evaluated with their released implementations and configurations.
We compare PASTA with TextCenGen and SAWNA, which directly address reserved-region preservation or object exclusion for downstream text placement.
Layout-guided placement methods mainly use spatial conditions to encourage objects to appear in specified regions, whereas our task uses the reserved-region mask to prevent salient foreground objects from appearing in those regions.
Therefore, our experimental comparisons focus on methods that share the same reserved-region objective.

\subsection{Dataset}
\label{ssec:d_and_m}

We conduct experiments on the P2P~\cite{hertz2022prompt} and SynPrompt~\cite{liang2024textcengen} datasets according to  TextCenGen~\cite{liang2024textcengen}, which are benchmarks for space-aware text-to-image generation, providing paired text prompts and reserved spatial masks.
In the original setup, each prompt is evaluated under a \emph{small-mask} setting using five reserved-region masks, 
whose average area ratio is approximately 3.8\%.

While this setting reflects lightweight text-placement scenarios, it does not capture practical design cases that require large clean regions.
To evaluate robustness under stricter and more realistic constraints, we extend the evaluation protocol by introducing an additional \emph{large-mask} setting.
Specifically, for each prompt, we introduce ten reserved-region masks with an average area ratio of approximately 30\%, simulating scenarios that require substantial negative space.

\subsection{Metrics}
We evaluate overall generation quality and text--image alignment using CLIP Score~\cite{radford2021learning} and HPSv2~\cite{wu2023human}.
The CLIP Score measures the semantic consistency between the generated image and the input prompt, where higher values indicate better alignment.
HPSv2 is a perceptual quality metric that aggregates human preference by jointly considering visual quality and semantic plausibility.
To specifically evaluate reserved-region fidelity, we introduce an \emph{intrusion rate}.
For each generated image, we first perform object detection with YOLO-World~\cite{cheng2024yolo} with the foreground object name specified in the prompt as a text query.
Detected bounding boxes are then processed by SAM2~\cite{ravi2024sam} to obtain pixel-level foreground masks, which are aggregated across candidate masks and bounding boxes into a single foreground mask.
We compare this aggregated mask with the predefined reserved-region mask, where white regions indicate areas where objects should not appear.
The intrusion rate is defined as the proportion of foreground pixels falling inside the reserved region, with lower values indicating better suppression. When YOLO-World detects no foreground objects, we exclude the sample from the intrusion-rate computation.
In our evaluation, such cases are rare and typically occur when the foreground object is extremely small or visually indistinguishable from the background.
Since these cases do not introduce visually salient content into the reserved region and do not interfere with downstream text or logo placement, excluding them yields a more faithful assessment of practical reserved-region violations.

\subsection{Qualitative Results}
\label{ssec:qualitative}
\begin{figure*}[t]
    \centering
    \includegraphics[width=\linewidth]{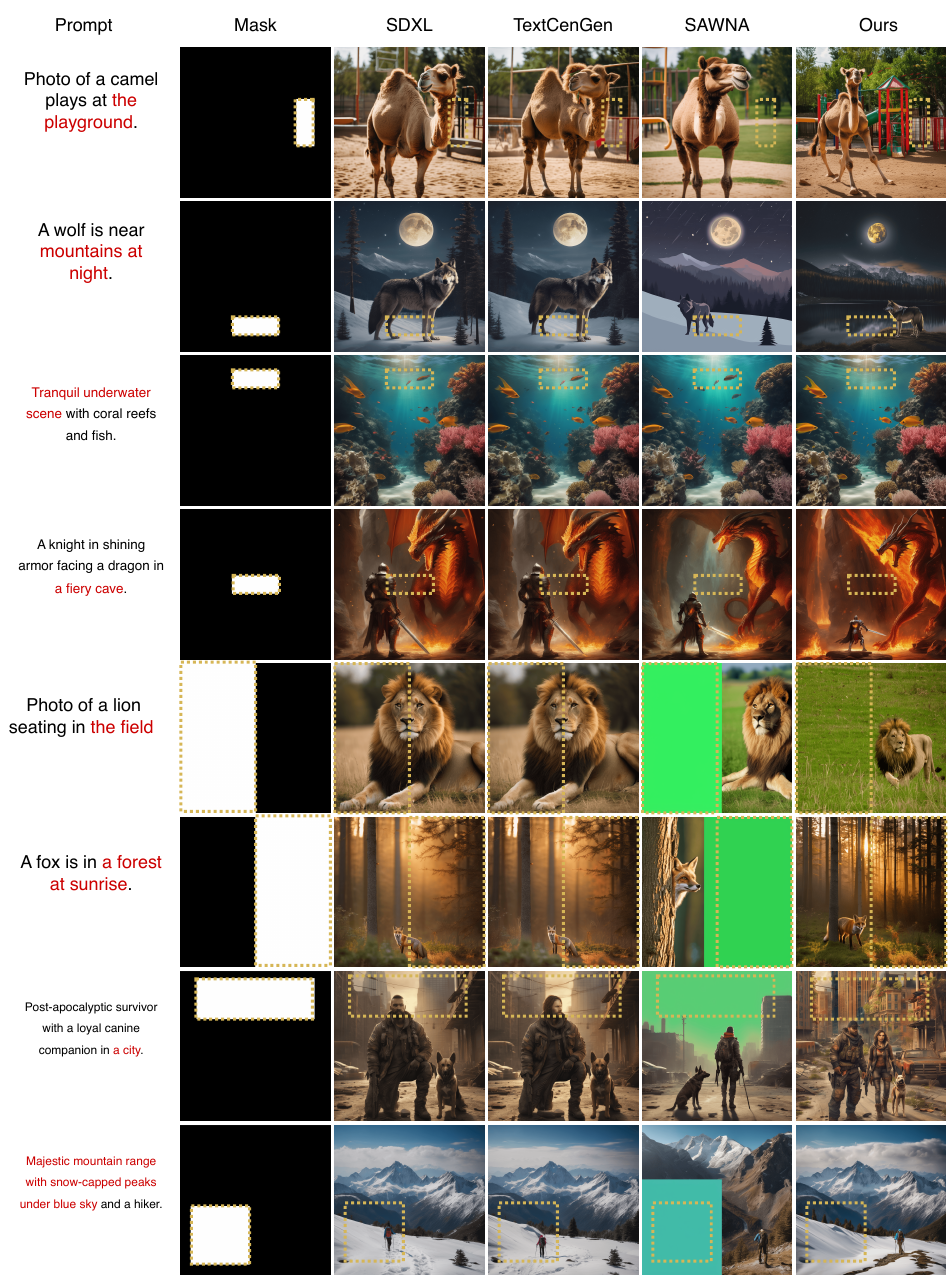}
    \caption{
    Qualitative comparison under small- and large-mask settings.
Rows 1--4 show small-mask examples, while Rows 5--8 show large-mask examples.
Within each mask setting, the first two rows show SynPrompt examples and the last two rows show P2P examples.
Red words in the prompt column indicate the context prompt \(P_{\mathrm{ctx}}\), and yellow dashed boxes denote reserved regions.
TextCenGen can slightly adjust object placement under small-mask settings but often fails to satisfy large-mask constraints.
SAWNA suppresses object generation inside the reserved regions, but frequently produces flat or near-monochrome backgrounds with visible discontinuities.
PASTA preserves clean reserved regions while maintaining natural and coherent backgrounds across both mask sizes.
}    
    \label{fig:result}
\end{figure*}

Fig.~\ref{fig:result} shows qualitative comparisons on SynPrompt and P2P prompts under both small- and large-mask settings.
Rows 1--4 show small-mask examples, while Rows 5--8 show large-mask examples.
Within each mask setting, the first two rows show SynPrompt examples and the last two rows show P2P examples.
In this figure, the red words in the prompt column indicate the context prompt \(P_{\mathrm{ctx}}\), and the yellow dashed boxes denote the reserved regions.
TextCenGen removes or slightly relocates objects when the reserved region is small.
However, the visual differences are often subtle, and the main object frequently remains close to the target area.
When the reserved region becomes large, TextCenGen may fail to sufficiently displace the object and may not fully satisfy the spatial constraint, resulting in violations of the reserved-region requirement.
SAWNA prevents the generation of objects within the reserved region under small masks.
However, injecting non-reactive noise into the masked area tends to cause the background in that region to collapse into flat or near-monochrome patterns.
This effect becomes more pronounced when the surrounding scene contains rich textures or complex colors, leading to noticeable visual discontinuities between reserved regions and object-allowed regions, especially for large masks.
In contrast, PASTA consistently follows the reserved-region constraint across both small and large masks.

\FloatBarrier

\subsection{Quantitative Results}
\label{ssec:quantitative}
Table~\ref{tab:quantitative_p2p_llm_small_large} summarizes quantitative results on P2P and SynPrompt prompts under small and large reserved-region masks.
PASTA consistently achieves substantially lower intrusion rates than SDXL and TextCenGen under the large-mask setting, while maintaining higher CLIP and HPS scores than SAWNA.
In contrast, TextCenGen shows higher intrusion rates than PASTA under large masks, and the vanilla SDXL baseline violates the spatial constraint.
Under the small-mask setting, PASTA remains competitive with existing space-aware methods in terms of CLIP and HPS scores.
TextCenGen achieves higher CLIP and HPS scores than PASTA under the large-mask setting on both datasets, while PASTA achieves substantially lower intrusion rates.
SAWNA yields the lowest intrusion rates under large masks but shows lower CLIP and HPS scores.
Overall, PASTA provides a favorable balance between reserved-region enforcement and generation quality, substantially reducing intrusion under large masks while maintaining competitive text-image alignment and perceptual quality across both P2P and SynPrompt prompts.

\begin{table}[t]
\centering
\caption{Quantitative comparison under small and large reserved-region masks.
Higher is better for CLIP and HPS, while lower is better for Intrusion Rate.}
\label{tab:quantitative_p2p_llm_small_large}

\footnotesize
\renewcommand{\arraystretch}{1.15}
\setlength{\tabcolsep}{6pt}

\begin{tabular}{llccc}
\toprule
Method & Setting & CLIP$\uparrow$ & HPS$\uparrow$ & Intr.$\downarrow$ \\
\midrule

\multicolumn{5}{c}{\textbf{P2P Prompts}} \\
\midrule

SDXL & Small & 33.28 & 0.291 & 0.278 \\
TextCenGen~\cite{liang2024textcengen}
& Small & 33.06 & \textbf{0.291} & 0.287 \\
SAWNA~\cite{morita2025sawna}
& Small & \textbf{33.25} & \underline{0.284} & \underline{0.148} \\
Ours
& Small & \underline{33.08} & \underline{0.284} & \textbf{0.098} \\

\cmidrule(lr){1-5}

SDXL & Large & 33.28 & 0.291 & 0.223 \\
TextCenGen~\cite{liang2024textcengen}
& Large & \textbf{33.02} & \textbf{0.290} & 0.222 \\
SAWNA~\cite{morita2025sawna}
& Large & 31.21 & 0.273 & \textbf{0.014} \\
Ours
& Large & \underline{32.26} & \underline{0.277} & \underline{0.038} \\

\midrule

\multicolumn{5}{c}{\textbf{SynPrompt Prompts}} \\
\midrule

SDXL & Small & 33.73 & 0.293 & 0.119 \\
TextCenGen~\cite{liang2024textcengen}
& Small & 33.63 & \underline{0.286} & 0.110 \\
SAWNA~\cite{morita2025sawna}
& Small & \underline{33.67} & \textbf{0.287} & \textbf{0.073} \\
Ours
& Small & \textbf{33.93} & 0.284 & \underline{0.076} \\

\cmidrule(lr){1-5}

SDXL & Large & 33.73 & 0.293 & 0.104 \\
TextCenGen~\cite{liang2024textcengen}
& Large & \textbf{33.65} & \textbf{0.284} & 0.101 \\
SAWNA~\cite{morita2025sawna}
& Large & 31.97 & 0.240 & \textbf{0.013} \\
Ours
& Large & \underline{33.59} & \underline{0.281} & \underline{0.048} \\

\bottomrule
\end{tabular}
\end{table}

\subsection{Ablation Study}
\label{ssec:ablation}

We conduct an ablation study to analyze the effect of the annealed mixing weight $\alpha$ on P2P prompts and the large-mask setting, highlighting differences in reserved-region fidelity.

\begin{figure*}[t]
    \centering
    \includegraphics[width=\textwidth]{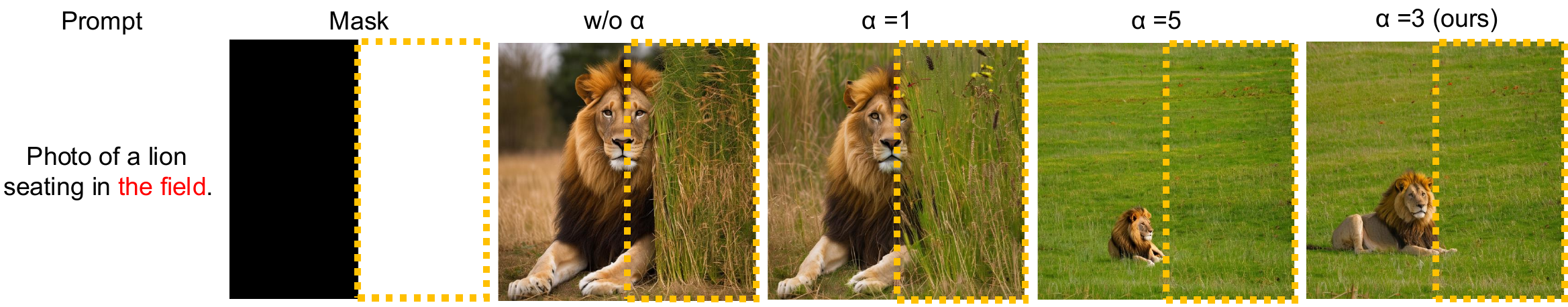}
    \caption{
Qualitative ablation of the annealed blending schedule.
All results use the same prompt, mask, and random seed. Red words in the prompt column indicate the context prompt \(P_{\mathrm{ctx}}\), and yellow dashed boxes denote reserved regions.
Without \(\alpha\) or with a weak constant weight \(\alpha=1\), distracting foreground or vegetation-like structures remain inside the reserved region.
A strong constant weight \(\alpha=5\) produces a cleaner region but over-suppresses the foreground object.
The proposed annealed schedule better balances reserved-region cleanliness and foreground fidelity.
}
    \label{fig:ablation_alpha}
\end{figure*}

Table~\ref{tab:ablation_alpha} compares our full method with a variant that removes the annealed weight schedule.
Without $\alpha$, the model achieves slightly higher CLIP and HPS scores, indicating stronger text alignment and perceptual quality.
However, this comes at the cost of a substantially higher intrusion rate, leading to foreground objects frequently leaking into the reserved region.
In contrast, introducing the annealed $\alpha$ dramatically reduces the intrusion rate,
demonstrating the importance of time-dependent guidance control for enforcing clean reserved regions.
Although CLIP and HPS scores decrease slightly, the model remains competitive while achieving better spatial fidelity.
These results confirm that the annealed weight schedule plays a crucial role in balancing semantic preservation and spatial constraint enforcement, and is essential for robust performance under large reserved-mask settings.
To further clarify the visual effect of the annealed schedule, Fig.~\ref{fig:ablation_alpha} compares different choices of \(\alpha\) using the same prompt, reserved-region mask, and random seed.
The qualitative results are consistent with Table~\ref{tab:ablation_alpha}.
Without \(\alpha\), the foreground object clearly intrudes into the reserved region, and distracting vegetation-like structures remain inside the mask.
A weak constant weight \(\alpha=1\) is still insufficient to produce a clean reserved region, while a strong constant weight \(\alpha=5\) suppresses the intrusion more aggressively but also over-suppresses the foreground object, producing it at an undesirably small scale.
In contrast, the proposed annealed schedule preserves a cleaner reserved region while maintaining an appropriate foreground-object size.
This indicates that the schedule does not simply increase suppression strength, but balances reserved-region cleanliness and prompt-faithful foreground generation.

\begin{table*}[t]
\centering
\captionsetup{width=\linewidth}
\caption{Ablation study on the effect of the annealed weight $\alpha$ on reserved-region preservation. Experiments are conducted using P2P prompts and the large-mask setting.
Higher is better for CLIP and HPS, while lower is better for Intrusion Rate.}
\label{tab:ablation_alpha}
\renewcommand{\arraystretch}{1.2}
\setlength{\tabcolsep}{6pt}
\begin{tabular}{l|ccc}
\toprule
\textbf{Setting} & \textbf{CLIP $\uparrow$} & \textbf{HPS $\uparrow$} & \textbf{Intrusion $\downarrow$} \\
\midrule
Ours (w/o annealed weight $\alpha$) & \textbf{33.035} & \textbf{0.2899} & 0.130 \\
Ours (w/ annealed weight $\alpha$)  & 32.263 & 0.2770 & \textbf{0.038} \\
\bottomrule
\end{tabular}
\end{table*}

\subsection{Limitation \& Future work}
\label{ssec:limitation}
While PASTA effectively handles a wide range of space-aware generation scenarios, it has limitations under extreme spatial constraints.
In particular, when the reserved region occupies a very large portion of the image (e.g., more than 80\%) or when the remaining generation area becomes extremely narrow, the model may fail to produce plausible content.
In such cases, the denoising process struggles to reconcile the strong spatial restriction with the learned image priors.

We hypothesize that these failure cases arise because such configurations are underrepresented or absent in the training data of large-scale text-to-image models.
Addressing these extreme scenarios may require explicitly curated training data or fine-tuning strategies tailored to highly constrained layouts.
We leave the exploration of data-driven or adaptive training approaches for handling these rare but practically relevant cases as future work.

\section{Conclusion}
\label{sec:conclusion}
We proposed PASTA, a training-free diffusion sampler for space-aware text-to-image generation.
By operating on a shared latent trajectory and selectively fusing semantic guidance with an annealed schedule, PASTA enforces reserved-region constraints while preserving background naturalness.
Experiments on both P2P and SynPrompt show that PASTA significantly improves reserved-region fidelity under large-mask settings while remaining competitive in text alignment and perceptual quality.
These results suggest that PASTA provides a practical solution for design scenarios requiring clean regions for text or interface elements.

\section*{Acknowledgements}
This research has been supported by the Kayamori Foundation of Informational Science Advancement.

\bibliographystyle{splncs04}
\bibliography{refs}

\end{document}